\documentclass{article}

\usepackage[preprint]{neurips_2026}

\usepackage[utf8]{inputenc} 
\usepackage[T1]{fontenc}    
\usepackage{hyperref}       
\usepackage{url}            
\usepackage{booktabs}       
\usepackage{amsfonts}       
\usepackage{nicefrac}       
\usepackage{microtype}      
\usepackage{xcolor}         

\usepackage{subcaption}
\usepackage[table]{xcolor}
\usepackage{graphicx}
\usepackage{multirow}
\usepackage{amsmath}
\definecolor{tablebg}{HTML}{F3F6FA}
\definecolor{tableline}{HTML}{D6DEE8}
\definecolor{larmbg}{HTML}{FFF3D6} 
\usepackage{float}
\usepackage[normalem]{ulem}

\newcommand{\vx}{\mathbf{x}}
\newcommand{\vy}{\mathbf{y}}

\title{Test-Time Compute Scaling for ASR with Depth-Conditioned Looped Transformers}

\author{
    Yacouba Kaloga$^{1,}$\thanks{Equal contribution.} \quad Shashi Kumar$^{1, 2, *}$ \quad Shakeel A. Sheikh$^{1*}$ \quad \textbf{Driss Khalil}$^{1}$ \\ \textbf{Petr Motlicek}$^{1, 3}$ \quad \textbf{Ina Kodrasi}$^{1}$\\ \\ 
    $^{1}$Idiap Research Institute, Switzerland \\
    $^{2}$EPFL, Switzerland \quad
    $^{3}$BUT, Czech Republic \\
    \texttt{\{yacouba.kaloga, shashi.kumar\}@idiap.ch} \\
    \texttt{shakeelzmail608@gmail.com}
}

\begin{document}

\maketitle

\begin{abstract}
End-to-end ASR systems typically use fixed-depth acoustic encoders at inference,
making it difficult to trade additional test-time computation for improved
recognition without training a larger model.
A natural approach is to
reuse a shared Transformer block recurrently, but we find that naive looping does not fully exploit additional recurrent compute. We introduce LARM, a depth-conditioned
looped Transformer that turns recurrent encoder depth into a controllable
test-time compute axis. LARM combines sparse CTC checkpoints,
supervision-clock embeddings, FiLM depth conditioning, and delayed
soft-posterior feedback. These components structure the loop into
recognition checkpoints separated by latent refinement phases and allow
shared weights to specialize across recurrent steps. On LibriSpeech, LARM
improves WER as the number of inference loops increases and achieves
performance competitive with deeper unshared-parameter baselines. Our results show that test-time
compute scaling can extend beyond autoregressive language-model reasoning
to continuous non-autoregressive speech recognition.
\end{abstract}

The code and selected checkpoints will be released soon. In the meantime, please contact
the authors if you need access or have any questions.

\section{Introduction}
End-to-end automatic speech recognition (ASR) models typically use a fixed-depth acoustic encoder followed by a CTC, transducer, or attention-based prediction head.
While this design has been effective, the computational depth of the model is fixed by the architecture and remains unchanged at inference.
This limits the ability of ASR models to benefit from additional test-time computation, which could in principle enable further contextualization, refinement of uncertain acoustic evidence, and improved predictions without changing the learned parameters. In practice, increasing inference computation therefore typically requires deploying a deeper or larger model, changing the parameterization and requiring separate training. Unlike architectures that support extended iterative computation, standard fixed-depth encoders offer no natural way to spend additional inference compute on progressively refining acoustic representations.

Recently, test-time compute scaling has gained significant traction in reasoning and language modeling, where autoregressive generation naturally supports longer inference-time computation that can act as a powerful substitute for parameter scaling~\citep{graves2016adaptive, wang2025hierarchical}. This has renewed interest in architectures that decouple computational depth from parameter count. Looped or recursive Transformers, in which a shared block of weights is applied repeatedly, offer a natural mechanism for such decoupling~\citep{univt, giannou2023looped, jolicoeur2025less}. However, adapting this paradigm to continuous, non-autoregressive sequence-to-sequence tasks like ASR is non-trivial. While iterative refinement in ASR has been explored, prior works primarily focus on masking or correcting discrete output hypotheses~\citep{higuchi20b_interspeech, Chi2021} rather than progressively deepening the core acoustic-linguistic representation. Motivated by recent work on looped architectures, a natural alternative is to apply recurrence directly within the acoustic encoder, reusing the same block across multiple refinement steps. In our early experiments, however, we found that simply reapplying the same encoder block was not sufficient to obtain competitive ASR performance, motivating additional conditioning mechanisms that allow successive iterations to specialize.

\begin{figure}[t]
    \centering
    \begin{minipage}{0.44\linewidth}
        \centering
        \includegraphics[width=\linewidth]{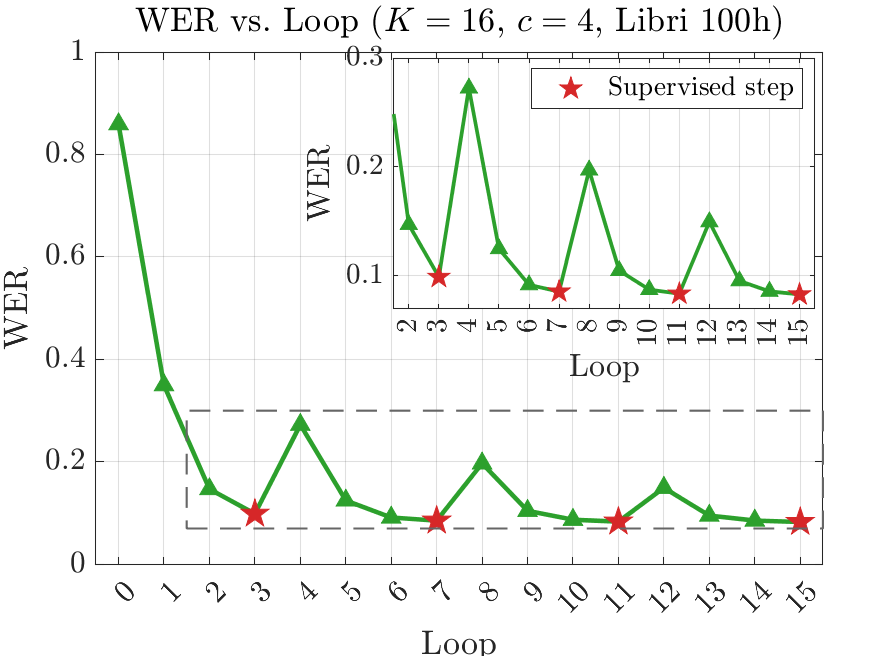}
        \vspace{-0.8em}
        \centerline{\small (a) WER across recurrent loops}
    \end{minipage}
    \hfill
    \begin{minipage}{0.44\linewidth}
        \centering
        \includegraphics[width=\linewidth]{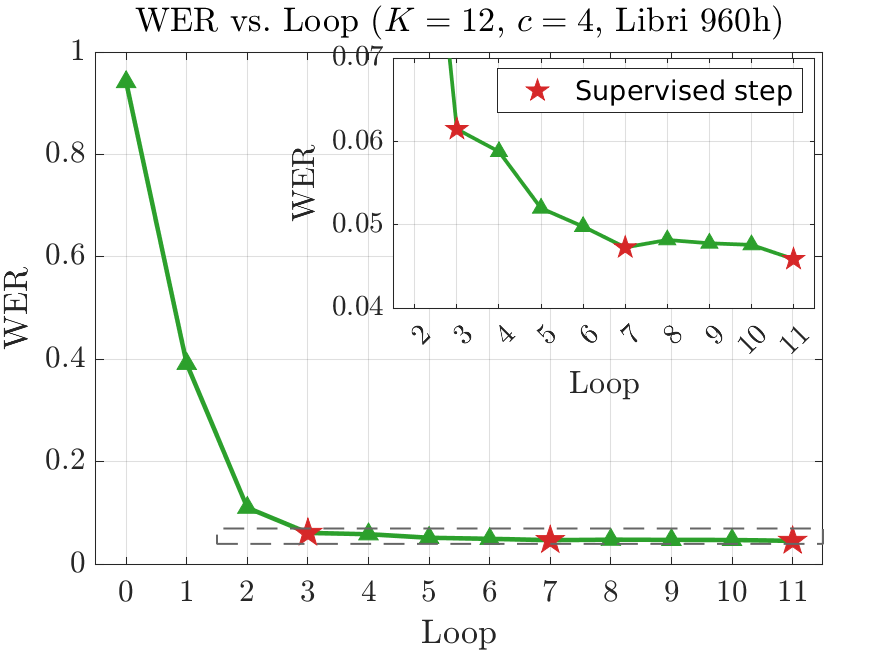}
       \centerline{\small (b) Smooth refinement regime}
    \end{minipage}

\caption{
\textbf{Loop-structured test-time computation in LARM.}
WER trajectories for two trained LARM models on LibriSpeech 100h
($K=16$, $c=4$) and 960h ($K=12$, $c=4$). Red stars denote supervised
loop. The trajectories show two refinement regimes:
one with non-monotonic intermediate loops and one with smoother
improvement. In both cases, supervised checkpoints improve from one
checkpoint to the next, supporting the role of sparse supervision as an
anchor for latent acoustic refinement.
}
    \label{fig:larm_overview}
    \vspace{-0.5cm}
\end{figure}

\begin{figure}[t]
    \centering
    \includegraphics[width=0.9\linewidth]{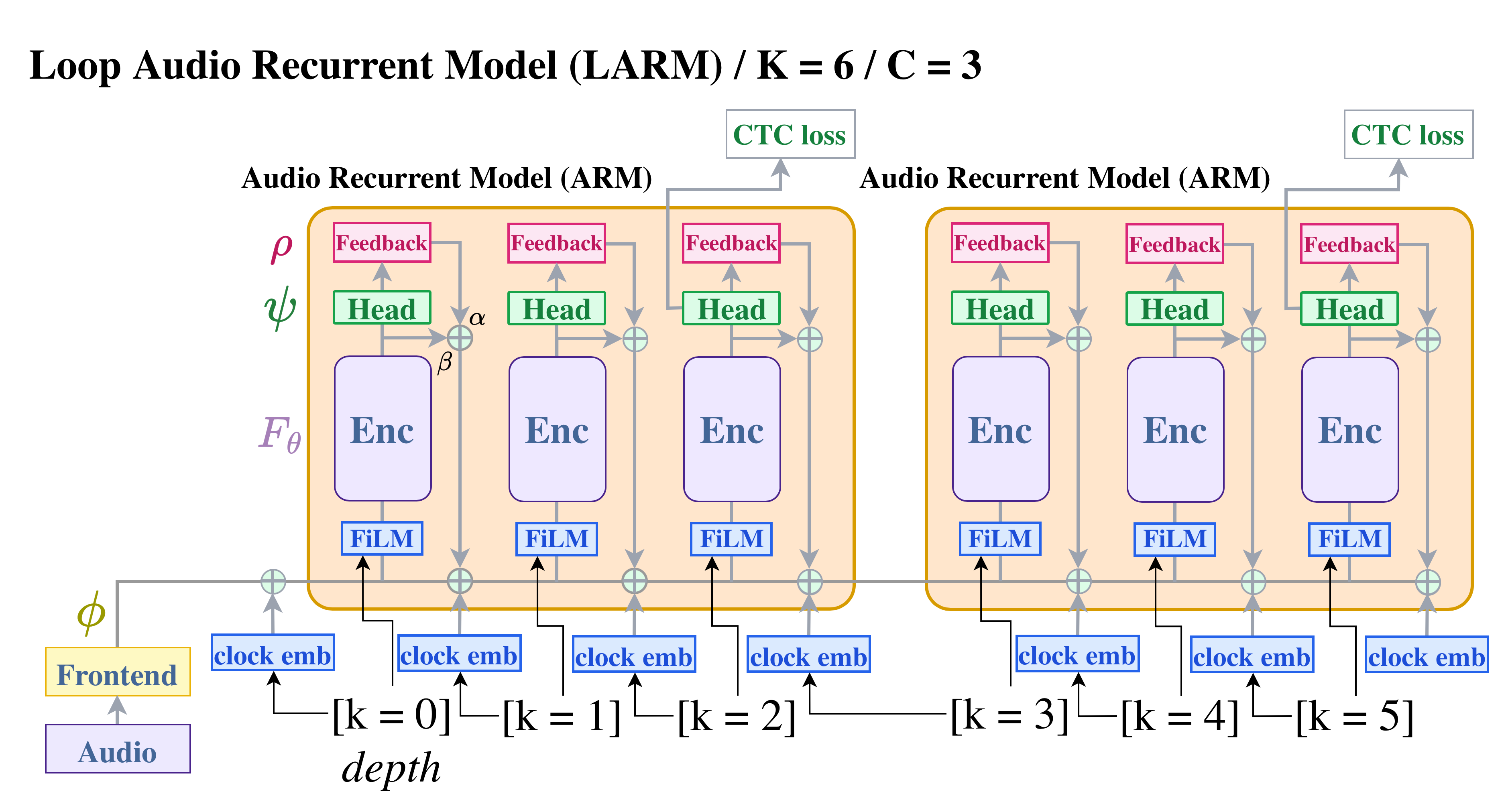}
    \caption{
    \textbf{Overview of the LARM architecture.}
    The acoustic frontend maps the input audio into an initial latent
    representation, which is processed recurrently by a shared encoder
    $F_{\theta}$. Each loop applies the same encoder parameters, a shared
    CTC head $\psi$, FiLM depth conditioning based on the normalized loop
    index $k$, supervision-clock embeddings based on $k \bmod c$, and
    delayed prediction feedback. CTC loss is applied only at sparse
    recognition checkpoints, while the intervening loops refine the latent
    acoustic representation before the next supervised checkpoint. The
    learnable scalars $\alpha$ and $\beta$ control the contribution of
    prediction feedback. The input features processed by the frontend are passed to each loop. The complete architecture is described in Section~\ref{sec:method}.
    }
    \label{fig:larm_architecture}
\end{figure}

To enable encoder-level test-time scaling for ASR, we propose the \textbf{Loop Audio Recurrent Model (LARM)}, a shared-parameter architecture that conditions repeated computation to support stage-specific processing.  LARM applies a shared Transformer block recurrently to a latent acoustic sequence, effectively treating computational depth as a dynamic axis. To avoid redundant computation across recurrent steps, LARM introduces three key structural mechanisms (see Figure~\ref{fig:larm_architecture}). First, we employ a sparse supervision schedule and a supervision clock embedding. By applying CTC loss only at periodic loop intervals, we structure the iterative process into a sequence of supervised recognition checkpoints separated by intermediate refinement steps that are not directly supervised by CTC (see Figure~\ref{fig:larm_overview}). Second, we utilize FiLM-based depth conditioning \citep{perez2018film}, which explicitly modulates the hidden state based on the normalized loop iteration, allowing the shared weights to perform specialized roles as computation progresses. Finally, we introduce delayed prediction feedback, an ASR-specific recurrent mechanism where soft CTC posteriors from the previous iteration are shifted by one frame and reinjected into the network. This explicitly propagates left-to-right token-level continuity through the depth of the model, guiding the trajectory of acoustic refinement. 

Our main contributions can be summarized as follows:
\begin{itemize}
    \item \textbf{Encoder-Level Test-Time Scaling for ASR:} We introduce LARM, a depth-conditioned looped Transformer for ASR that decouples parameter count from inference depth, enabling dynamic test-time compute scaling.
    \item \textbf{Stage-Specialized Recurrent Training:} We propose a novel loop-structuring mechanism combining sparse intermediate CTC supervision, periodic clock embeddings, and FiLM depth conditioning, successfully specializing a shared encoder across recurrent iterations.
    \item \textbf{Delayed Prediction Feedback:} We introduce an ASR-specific feedback mechanism that reinjects shifted soft-token posteriors into the recurrent state, improving iterative acoustic recognition.
    \item \textbf{Empirical validation on LibriSpeech~\citep{panayotov2015librispeech}:} We show that LARM is an effective looped architecture for ASR, improving as the number of loops increases, while its structural mechanisms make each loop more effective and achieve strong performance with fewer loops. LARM achieves performance competitive with deeper, unshared-parameter baselines while using a fraction of the parameter count.
\end{itemize}

\section{Related Work}

We position LARM at the intersection of test-time compute scaling, looped Transformer architectures, and iterative refinement methods for ASR.

\paragraph{Test-Time Compute Scaling.}
Dynamically allocating additional computation at inference has long
been studied, with Adaptive Computation Time providing an early mechanism
for input-dependent recurrent computation~\citep{graves2016adaptive}.
More recently, test-time compute scaling has become central in large
language models, where search, verification, self-correction, and
iterative refinement can substantially improve reasoning performance
without changing model parameters~\citep{snell2024scaling,
jaech2024openai, muennighoff2025s1, madaan2023self}. A related direction
is latent reasoning, where models perform additional internal computation
before emitting outputs, for example through pause tokens, internal
rationales, or filler-token computation~\citep{goyal2023think,
zelikman2024quiet, pfau2024let, wang2025hierarchical}. These methods
primarily exploit the autoregressive structure of language generation,
whereas our goal is to expose a similar compute-scaling axis inside a
continuous ASR encoder.

\paragraph{Looped Architectures.}

Architectures that decouple computational depth from parameter count
provide a natural mechanism for this goal. Cross-layer parameter sharing
was used in ALBERT to reduce memory and regularize deep models
\citep{lan2019albert}, while the Universal Transformer applied shared
Transformer blocks recurrently across depth~\citep{univt}. Related
formulations include Deep Equilibrium Models, which repeatedly apply a
shared transformation to solve for a fixed point~\citep{bai2019deep}.
Recent work further shows that recurrent or looped Transformers can act
as powerful iterative computers, with both theoretical and empirical
evidence for improved algorithmic and reasoning behavior
\citep{perez2019turing, giannou2023looped, schwarzschild2021can,
wang2025hierarchical, jolicoeur2025less}. LARM adapts this looped
Transformer view to acoustic modeling, using recurrent depth as a
test-time compute axis rather than only as a parameter-sharing device.

\paragraph{Iterative Refinement and Feedback in ASR.}

In ASR, iterative refinement has mainly been used to narrow the gap
between non-autoregressive (NAR) and autoregressive decoding. Methods such
as Mask CTC~\citep{higuchi20b_interspeech}, Imputer~\citep{chan2020imputer},
and Align-Refine~\citep{Chi2021} repeatedly mask, correct, or realign
discrete textual hypotheses after an initial CTC-style prediction. While
effective, these approaches primarily refine outputs rather than deepen
the continuous acoustic representation. Closer to LARM are methods that
use intermediate acoustic predictions inside the encoder. Auxiliary
intermediate CTC losses improve the optimization of deep ASR encoders
\citep{sanabria2018hierarchical}, while Self-Conditioned CTC
\citep{nozaki2021relaxing} and Intermediate CTC~\citep{komatsu22_interspeech}
feed internal CTC predictions forward to condition later layers and relax
CTC's conditional-independence assumption. Self-Conditioned Folded
Encoders~\citep{9746770} similarly explore reusing acoustic blocks across
depth. However, these methods typically rely on fixed-depth computation,
distinct layer weights, dense intermediate supervision, or limited control
over the recurrent transition.

\paragraph{Positioning of Our Work.}

LARM brings the test-time compute scaling perspective of looped
architectures to continuous acoustic modeling. Unlike Mask CTC, Imputer,
and Align-Refine, which iteratively revise discrete output hypotheses,
LARM performs refinement inside the encoder by repeatedly transforming the
continuous acoustic-linguistic representation. Unlike self-conditioned CTC
models and folded encoders, which generally operate with fixed inference
depth, dense intermediate prediction losses, or limited control over how
computation changes across depth, LARM exposes recurrent encoder depth as
an explicit inference-time compute axis. It does so with a shared
Transformer block whose behavior is shaped by sparse recognition
checkpoints, supervision-clock embeddings, and FiLM depth conditioning.
Finally, LARM introduces delayed prediction feedback, reinjecting
one-frame-shifted soft CTC posteriors into the recurrent state so that
later loops can build on an evolving left-to-right token-level hypothesis.
Together, these mechanisms turn repeated encoder computation from simple
weight reuse into stage-specialized acoustic refinement, enabling
parameter-efficient test-time compute scaling for ASR.

\section{Method}

\label{sec:method}
In this section, we first introduce the standard looped Transformer
formulation, then present the Loop Audio Recurrent Model (LARM). We
describe its recurrent formulation, delayed prediction feedback, sparse
supervision, supervision clock embedding, and FiLM-based depth
conditioning, which together enable efficient test-time compute scaling
through a shared-parameter loop.

\subsection{Standard Looped Transformer Formulation}
In ASR, the model maps an acoustic input sequence
$\vx = \{\vx_i\}_{i=1}^{n}$ to an output token sequence
$\vy = \{\vy_j\}_{j=1}^{m}$. Standard encoder-based architectures
increase representational depth by stacking distinct layers, so depth
and parameter count typically grow together. A looped Transformer
instead applies a shared block $F_{\theta}$ repeatedly to a latent
representation, decoupling computational depth from parameter count.
Given an input sequence $\vx$, an initial latent representation is first
obtained using a feature encoder $\phi$, after which the same block
$F_{\theta}$ is applied for $K$ loops:
\begin{equation}
    \mathbf{h}^{(0)} = \phi(\vx),
    \qquad
    \tilde{\vy}^{(K)}
    =
    \psi\!\left(
    \underbrace{
    F_{\theta} \circ F_{\theta} \circ \cdots \circ F_{\theta}
    }_{K\ \text{times}}
    \left(\mathbf{h}^{(0)}\right)
    \right),
\end{equation}
where $\psi$ denotes the CTC head.

\subsection{Loop Audio Recurrent Model Architecture}

Building on this formulation, we describe how LARM structures recurrent
shared-parameter computation for acoustic recognition.
Figure~\ref{fig:larm_architecture} provides an overview of the model.

\subsubsection{Preliminaries}

LARM distinguishes the output of the shared encoder at loop $k$, denoted
$\mathbf{z}^{(k)}$, from the recurrent state passed to the next loop,
denoted $\mathbf{h}^{(k)}$. The recurrent state $\mathbf{h}^{(k)}$ is
formed by combining the encoder output with prediction feedback, the
frontend skip connection, and loop-conditioning signals.

\textbf{Acoustic Frontend.}
LARM maps log-Mel features $\vx \in \mathbb{R}^{T \times 80}$ to the
initial recurrent state $\mathbf{h}^{(0)} \in \mathbb{R}^{T' \times d}$
using an acoustic frontend $\phi$ with two strided 2D
convolutional subsampling layers, a linear projection to dimension $d$,
and dropout. The frontend reduces the time resolution by a factor of $4$.

\textbf{CTC Prediction Head.}
Token predictions are produced by a shared CTC head $\psi$ applied to an
encoder output $\mathbf{z}^{(k)} \in \mathbb{R}^{T' \times d}$. It maps
the representation to frame-level logits
$\boldsymbol{\ell}^{(k)} \in \mathbb{R}^{T' \times |\mathcal{V}|}$,
where $\mathcal{V}$ includes the CTC blank symbol:
\begin{equation}
    \boldsymbol{\ell}^{(k)}
    =
    \psi(\mathbf{z}^{(k)})
    =
    \mathbf{z}^{(k)} \mathbf{W}_{\psi}^{\top}
    +
    \mathbf{b}_{\psi},
    \qquad
    \mathbf{p}^{(k)}
    =
    \mathrm{softmax}\!\left(\boldsymbol{\ell}^{(k)}\right).
\end{equation}
Further implementation details are provided in Appendix~\ref{app:ctc_head}.

\textbf{Shared Encoder.}
The shared encoder $F_{\theta}$ is composed of $N$ stacked pre-norm
Transformer blocks. 
followed by a feed-forward network with residual connections.
The same encoder parameters are reused across all loops. 
Implementation details are provided in Appendix~\ref{app:shared_encoder}.

\subsubsection{Sparse Supervision and Loop Structure}
A looped encoder produces a sequence of encoder outputs
$\{\mathbf{z}^{(k)}\}_{k=1}^{K}$. Two natural supervision strategies are
to apply the loss either only at the final loop or densely at every loop.
Final-loop supervision leaves intermediate loop outputs without direct
recognition supervision, while dense supervision provides recognition
signals throughout the loop but encourages every loop to behave as a
complete recognition stage.

LARM adopts an intermediate regime based on sparse supervision. The goal
is to structure the loop as a sequence of recognition checkpoints
separated by intervals of intermediate computation. We refer to the
supervision period $c$ as the \emph{checkpoint interval}: it determines
the spacing between directly supervised recognition stages, leaving
$c-1$ intermediate loops for latent refinement between consecutive
checkpoints. Selected loops therefore act as recognition stages, while
intermediate loops remain available to refine the representation before
the next checkpoint. As illustrated in
Figure~\ref{fig:larm_architecture}, supervised checkpoint loops are
separated by intermediate recurrent refinement steps.

Given a checkpoint interval $c$ with $c \mid K$ ($c$ divide $K$), only a subset of loops
is directly optimized:
\begin{equation}
    \mathcal{S} = \{c, 2c, \dots, K\}.
\end{equation}
For each supervised loop $k \in \mathcal{S}$, the model produces CTC
logits $\boldsymbol{\ell}^{(k)} = \psi(\mathbf{z}^{(k)})$, and the
training objective is
\begin{equation}
    \mathcal{L}
    =
    \frac{1}{|\mathcal{S}|}
    \sum_{k \in \mathcal{S}}
    \mathcal{L}_{\mathrm{CTC}}
    \left(
        \boldsymbol{\ell}^{(k)}, \vy
    \right).
\end{equation}

\subsubsection{Prediction Feedback and State Aggregation}

At loop $k$, the shared encoder maps the current recurrent state
$\mathbf{h}^{(k-1)}$ to an updated representation
\begin{equation}
    \mathbf{z}^{(k)} = F_\theta(\mathbf{h}^{(k-1)}).
\end{equation}
The shared CTC head can then be applied to $\mathbf{z}^{(k)}$ to obtain
the posterior $\mathbf{p}^{(k)}$ as defined above, even when loop $k$ is
not directly supervised. These posteriors provide a soft token-level hypothesis that is reused as
recurrent feedback. We project them back to the hidden space using a
learned feedback projection $\rho$ :
\begin{equation}
    \mathbf{r}^{(k)}_t
    =
    \rho(\mathbf{p}^{(k)}_t)
    =
    \mathbf{p}^{(k)}_t \mathbf{W}_{\rho}.
\end{equation}

where $\mathbf{W}_{\rho} \in \mathbb{R}^{|\mathcal{V}| \times d}$. To inject left-context information, LARM applies a one-frame delay along
the time axis:
\begin{equation}
    \bar{\mathbf{r}}^{(k)}_t
    =
    \mathbf{r}^{(k)}_{t-1},
\end{equation}
with zero padding at $t=0$. The delayed feedback is then combined with
the encoder output and the fixed frontend representation to form an
aggregated state:
\begin{equation}
    \mathbf{a}^{(k)}
    =
    \mathbf{z}^{(k)}
    +
    \beta\,\mathbf{h}^{(0)}
    +
    \alpha\,\bar{\mathbf{r}}^{(k)},
\end{equation}
where $\mathbf{h}^{(0)}=\phi(\vx)$ and $\alpha,\beta$ are learnable
scalars. This aggregation lets each loop condition on both the original
acoustic representation and a delayed soft prediction signal. Across
loops, this propagates token-level continuity through recurrent depth and
encourages later predictions to follow more consistent local decoding
trajectories.

\subsubsection{Clock and Depth Conditioning}

Sparse supervision gives LARM a periodic loop structure, while repeated
shared computation also requires the model to distinguish early from late
loops. LARM therefore conditions each recurrent update using two
complementary signals: a supervision clock embedding and FiLM depth
conditioning.

\textbf{Supervision clock embedding.}  The supervision clock encodes the position of the current loop within
the checkpoint interval $c$. For loop $k=1,\dots,K$, LARM selects
\begin{equation}
    \mathbf{e}_{\mathrm{clock}}^{(k)}
    =
    \mathbf{W}_c[(k-1) \bmod c],
    \qquad
    \mathbf{W}_c \in \mathbb{R}^{c \times d}.
\end{equation}
The embedding is broadcast across time and added to the aggregated state:
\begin{equation}
    \hat{\mathbf{a}}^{(k)}_t
    =
    \mathbf{a}^{(k)}_t
    +
    \mathbf{e}_{\mathrm{clock}}^{(k)}.
\end{equation}

\textbf{FiLM depth conditioning.} In addition, LARM uses FiLM modulation to condition the recurrent update
on absolute loop depth. We define
\begin{equation}
    \bar{d}(k) = \frac{k-1}{K-1},
\end{equation}
and compute the next recurrent state as
\begin{equation}
    \mathbf{h}^{(k)}
    =
    \boldsymbol{\gamma}_{\mathrm{film}}(\bar{d}(k))
    \odot
    \hat{\mathbf{a}}^{(k)}
    +
    \boldsymbol{\beta}_{\mathrm{film}}(\bar{d}(k)),
\end{equation}
where $\boldsymbol{\gamma}_{\mathrm{film}},
\boldsymbol{\beta}_{\mathrm{film}} :
\mathbb{R}\to\mathbb{R}^{d}$ are small MLPs that produce feature-wise
scale and bias terms. This modulation lets the shared encoder implement
different transformations at different loop depths without untying its
parameters across loops.

The clock embedding makes the sparse supervision cycle explicit, while
FiLM depth conditioning allows the shared encoder to specialize across
early and late loops without untying its parameters.

\subsubsection{Overall Recurrent Procedure}

LARM applies the shared encoder for $K$ loops starting from the frontend
representation $\mathbf{h}^{(0)}=\phi(\vx)$. At loop $k$, the shared
encoder produces an updated representation $\mathbf{z}^{(k)}$ from the
previous recurrent state $\mathbf{h}^{(k-1)}$. The CTC prediction head
computes frame-level token posteriors, which are projected back to the
hidden space, shifted by one frame to form delayed prediction feedback,
and aggregated with both the encoder output and the initial acoustic
representation. The aggregated state is then augmented with the
supervision clock embedding and modulated by FiLM depth conditioning to
produce the next recurrent state $\mathbf{h}^{(k)}$.

During training, CTC supervision is applied only at the loops in
$\mathcal{S}$, defining a sequence of supervised recognition checkpoints
separated by intermediate refinement loops that are not directly
supervised. During inference, the CTC head can be applied at any
loop, including loops between supervised checkpoints. Overall, LARM forms
a shared-parameter recurrent encoder in which sparse supervision
structures the loop, delayed prediction feedback propagates token-level
context across recurrent depth, and loop conditioning encourages
successive loops to take distinct computational roles.

\section{Experimental Results}

\subsection{Experimental Setup}
\textbf{Data.}
We evaluate on the LibriSpeech benchmark~\citep{panayotov2015librispeech}.
Ablation experiments are conducted on the \texttt{train-clean-100}
split (100\,h), and scaling experiments use the full \texttt{train-960}
split (960\,h). Models are evaluated on \texttt{test-clean}, with
\texttt{test-other} additionally reported for the main LibriSpeech
results. Additional preprocessing details are provided in
Appendix~\ref{app:experimental_setup}.

\textbf{Architecture.}
LARM uses a convolutional subsampling frontend followed by a shared
Transformer encoder. The default configuration uses $N=4$ shared
Transformer blocks, model dimension $d=384$, and fixed per-head
dimension $d_h=64$, so the number of attention heads scales with width as
$H=d/d_h$. The recurrent loop is run for $K=12$ steps with checkpoint
interval $c=4$, yielding three supervised recognition checkpoints.
Prediction feedback uses the previous-frame mode, and the mixing scalars
$\alpha$ and $\beta$ are learned. Additional architecture details are
provided in Appendix~\ref{app:experimental_architecture}.

\textbf{Training and Decoding.}
Models are trained with AdamW~\citep{loshchilov2018decoupled}, gradient
clipping, and a cosine learning-rate schedule with linear warmup. Most
models are trained for 50 epochs, with extended-budget runs noted
separately. We use light SpecAugment~\citep{park2019specaugment} during
training. Additional optimization and augmentation details are provided in
Appendix~\ref{app:training_details}.
WER is reported using greedy CTC decoding unless specified otherwise.
For the main LibriSpeech benchmark results, we also report decoding with
a 4-gram KenLM model~\citep{heafield2011kenlm} using beam search
(beam width 100, LM weight 0.5, word insertion bonus 1.0).

\begin{table}[t]
\centering

\caption{LibriSpeech WER (\%) on \texttt{test-clean} and \texttt{test-other}. 
We report greedy CTC decoding and decoding with a 4-gram language model. 
 The 16-block encoder 
is included as a classical  conventional depth baseline. 
}
\label{tab:librispeech_main}
\renewcommand{\arraystretch}{1.15}
\setlength{\tabcolsep}{6pt}
\resizebox{\linewidth}{!}{%
\begin{tabular}{cc c@{\hspace{1.2em}}cccc}
\toprule
\multirow{2}{*}{\textbf{Training data}} 
& \multirow{2}{*}{\textbf{Model}}
& \multirow{2}{*}{\textbf{\#Params}}
& \multicolumn{2}{c}{\textbf{Greedy CTC}}
& \multicolumn{2}{c}{\textbf{+ 4-gram LM}} \\
\cmidrule(lr){4-5}
\cmidrule(lr){6-7}
& & & \textbf{\texttt{test-clean}} & \textbf{\texttt{test-other}}
& \textbf{\texttt{test-clean}} & \textbf{\texttt{test-other}} \\
\midrule

\multirow{3}{*}{\textbf{100h}}
& Standard encoder, 4 blocks   &  7.6M & 26.78 & 52.82 & 13.44 & 36.60 \\
& Standard encoder, 16 blocks  & 28.9M & 14.43 & 37.23 & 9.97  & 28.68 \\

&\cellcolor{larmbg} \textbf{LARM} (4 blocks, $K=12$) &  \cellcolor{larmbg} \textbf{7.7M} & \cellcolor{larmbg} \textbf{11.34} & \cellcolor{larmbg} \textbf{31.84} & \cellcolor{larmbg} \textbf{8.66} &  \cellcolor{larmbg}\textbf{26.28} \\

\midrule

\multirow{3}{*}{\textbf{960h}}
& Standard encoder, 4 blocks   &  7.6M & 14.07 & 29.82 & 6.77 & 17.39 \\
& Standard encoder, 16 blocks  & 28.9M &  4.79 & 13.26 & \textbf{3.51} & 9.87 \\

& \cellcolor{larmbg}\textbf{LARM} (4 blocks, $K=12$) &  \cellcolor{larmbg}\textbf{7.7M} & \cellcolor{larmbg} \textbf{4.59} & \cellcolor{larmbg} \textbf{11.75} & \cellcolor{larmbg} \textbf{3.51} & \cellcolor{larmbg} \textbf{9.38} \\

\bottomrule
\end{tabular}%
}
\vspace{-0.5cm}
\end{table}

\subsection{LibriSpeech Benchmark Results}

We first use controlled comparisons with compact Transformer encoders to
isolate whether recurrent shared computation provides a useful
test-time compute axis for ASR. We compare LARM (shaded row in
Table~\ref{tab:librispeech_main}) against standard non-looped encoders. Our reference LARM
model uses the smallest width considered in our experiments ($d=384$), a
shared encoder composed of $N=4$ Transformer blocks, and $K=12$ recurrent
loops with checkpoint interval $c=4$. We compare against two encoder baselines. The first uses the
same 4-block Transformer stack but executes it only once, approximately
matching the parameter count and isolating the effect of recurrent
computation. The second is a 16-block unshared encoder, representative
of a conventional deeper ASR encoder.

Table~\ref{tab:librispeech_main} shows that LARM substantially improves
over the parameter-matched non-looped encoder, indicating
that their is net gains coming from recurrent computation. On the 100h setting, LARM also outperforms the
16-block unshared encoder while using only 7.7M parameters, compared to
28.9M parameters. This indicates that shared recurrent depth can provide
a more parameter-efficient alternative to simply stacking additional
unshared layers.

The same trend carries over to the 960h setting. LARM improves over the
parameter-matched encoder by a large margin and slightly outperforms the
16-block unshared encoder under greedy decoding. With language-model
decoding, LARM matches the 16-block encoder on \texttt{test-clean} and
improves on \texttt{test-other}, indicating that the gains are not
restricted to the easier clean evaluation condition.

Overall, these results support the central claim of the paper: LARM
enables ASR models to trade additional test-time computation for improved
recognition without increasing the number of learned encoder parameters.

\subsection{Test-Time Compute Scaling}
\label{sec:test_time_compute_scaling}
We next ask whether a single trained LARM model can trade additional
inference computation for improved recognition. This differs from training
separate models with different loop budgets: here the parameters are
fixed, and only the number of executed recurrent loops changes. Figure~\ref{fig:larm_overview} shows that WER improves across supervised
checkpoints as more recurrent loops are executed. The intermediate loops
can behave differently across settings: the 100h model exhibits
non-monotonic refinement between checkpoints, while the 960h model follows
a smoother trajectory. In both cases, however, the supervised checkpoints
improve from one checkpoint to the next, showing that additional recurrent
computation can be converted into better recognition without changing the
learned parameters.

The checkpoint structure also provides natural early-exit points. For
example, in the 100h $K=16$ run, the third supervised checkpoint is
within 0.06 WER points of the final checkpoint, and the second checkpoint
is within 0.70 points. Thus, LARM can be 
stopped
early when a lower
latency inference is required. Detailed per-$K$ results are reported
in the appendix~\ref{app:detailed_test_time_scaling}.

\begin{table}[t]
\centering
\caption{Scaling LARM with model width on LibriSpeech. WER (\%) is reported on
\texttt{test-clean} and \texttt{test-other} using greedy CTC decoding and
decoding with a 4-gram language model. Bold indicates the best WER within
each training-data block; the shaded row is the reference small model.}
\label{tab:larm_width_scaling}
\renewcommand{\arraystretch}{1.15}
\setlength{\tabcolsep}{6pt}
\resizebox{\linewidth}{!}{%
\begin{tabular}{cc c@{\hspace{1.2em}}cccc}
\toprule
\multirow{2}{*}{\textbf{Training data}} 
& \multirow{2}{*}{\textbf{Model}}
& \multirow{2}{*}{\textbf{\#Params}}
& \multicolumn{2}{c}{\textbf{Greedy CTC}}
& \multicolumn{2}{c}{\textbf{+ 4-gram LM}} \\
\cmidrule(lr){4-5}
\cmidrule(lr){6-7}
& & & \textbf{\texttt{test-clean}} & \textbf{\texttt{test-other}}
& \textbf{\texttt{test-clean}} & \textbf{\texttt{test-other}} \\
\midrule
\multirow{4}{*}{\textbf{100h}}
& Standard encoder, 48 blocks  & 85.7M & 12.24 & 34.06 & 9.03  & 27.07 \\
& \cellcolor{larmbg} LARM ($K=12$, $d=384$)  & \cellcolor{larmbg}7.7M  & \cellcolor{larmbg}11.34 & \cellcolor{larmbg}31.84 &\cellcolor{larmbg} 8.66 &\cellcolor{larmbg} 26.28 \\
& LARM ($K=12$, $d=768$)  & 28.9M & 9.58  & 28.25 & 7.57 & 23.89 \\
& LARM ($K=12$, $d=1024$) & 85.7M & \textbf{8.95} & \textbf{27.93} & \textbf{7.22} & \textbf{23.79} \\

\midrule

\multirow{4}{*}{\textbf{960h}}& Standard encoder, 48 blocks  & 85.7M &  3.87   & 10.58    & 3.20   & 8.56 \\
&\cellcolor{larmbg} LARM ($K=12$, $d=384$)  &\cellcolor{larmbg} 7.7M  &\cellcolor{larmbg} 4.59 & \cellcolor{larmbg}11.75 & \cellcolor{larmbg}3.51 &\cellcolor{larmbg} 9.38 \\
& LARM ($K=12$, $d=768$)  & 28.9M & 3.45 & 9.44  & 2.93 & 7.93 \\
& LARM ($K=12$, $d=1024$) & 85.7M & \textbf{3.39} & \textbf{9.01} & \textbf{2.83} & \textbf{7.67} \\

\bottomrule
\end{tabular}%
}
\vspace{-0.3cm}
\end{table}
\subsection{Scaling with Model Size and Data}
We next evaluate whether LARM also benefits from encoder model and data
scaling. We vary the model dimension $d$, and compare performance on
LibriSpeech 100h and 960h. Table~\ref{tab:larm_width_scaling} shows that
increasing model width consistently improves LARM. We also include a
48-block standard encoder as a reference baseline at the largest scale:
this model matches the $d=1024$ LARM configuration both in parameter
count and in total number of Transformer-block applications, since the
reference LARM model uses a shared encoder with $N=4$ blocks and
$K=12$ loop iterations, for a total of $4\times 12$ block passes.

On LibriSpeech 100h, scaling from $d=384$ to $d=1024$ reduces greedy WER
from 11.34 to 8.95 on \texttt{test-clean} and from 31.84 to 27.93 on
\texttt{test-other}. The same trend holds with language-model decoding,
where WER decreases from 8.66 to 7.22 on \texttt{test-clean} and from
26.28 to 23.79 on \texttt{test-other}. Relative to the 48-block standard
encoder, the largest LARM model remains competitive while preserving the
shared-parameter recurrent structure.

Scaling the training data from 100h to 960h yields an even larger gain.
For the reference $d=384$ model, greedy WER improves from 11.34 to 4.59
on \texttt{test-clean} and from 31.84 to 11.75 on \texttt{test-other}.
Combining larger data with wider encoders further improves performance,
reaching 3.39 and 9.01 WER on \texttt{test-clean} and
\texttt{test-other}, respectively, for $d=1024$.

The gains from width become smaller at the largest scale, especially on
960h, where moving from $d=768$ to $d=1024$ gives only modest
improvements. This suggests that the current configuration begins to
saturate with respect to width alone. Additional scaling axes, including
longer training, larger loop budgets, and deeper shared Transformer
blocks, are reported in Appendix~\ref{app:extended_results}.

\begin{figure}[t]
    \centering

    \begin{subfigure}[t]{0.45\linewidth}
        \centering
        \includegraphics[width=\linewidth]{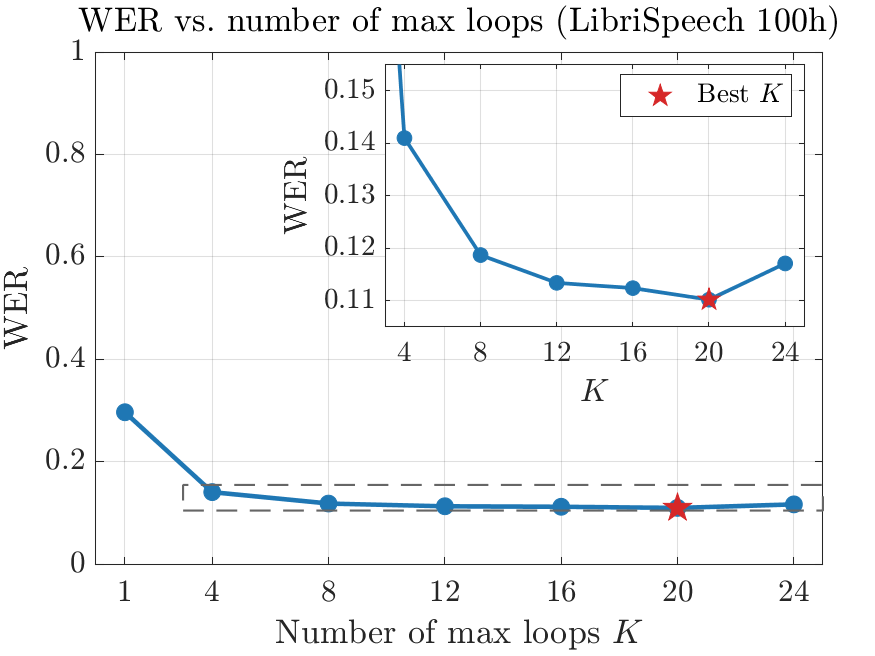}
        \caption{Trained loop-budget scaling}
        \label{fig:loop_budget_scaling}
    \end{subfigure}
    \hfill
    \begin{subfigure}[t]{0.45\linewidth}
        \centering
        \includegraphics[width=\linewidth]{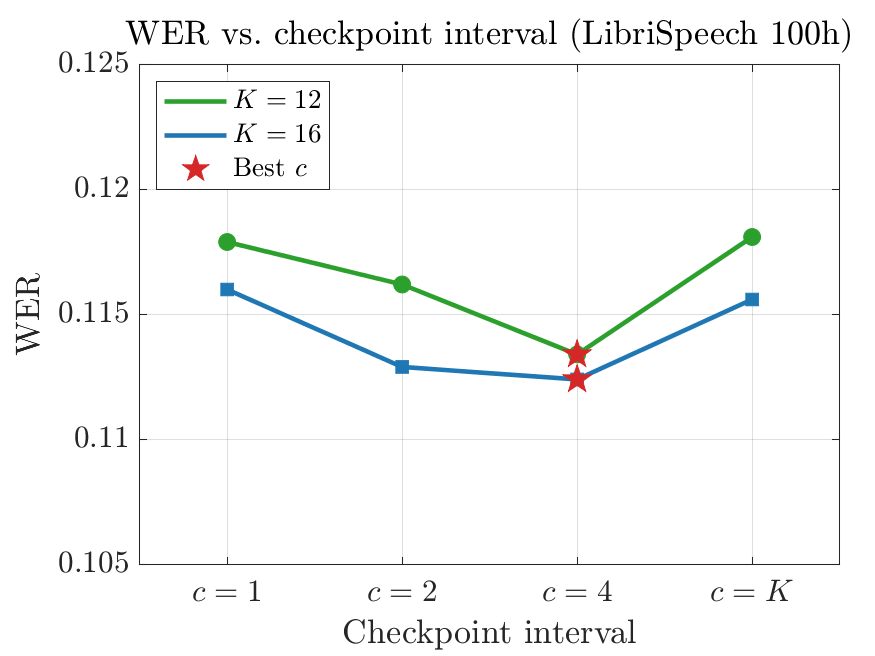}
        \caption{Sparse supervision schedule}
        \label{fig:checkpoint_interval_ablation}
    \end{subfigure}
\caption{
    \textbf{Training-time loop-structure ablations.}
    \subref{fig:loop_budget_scaling} Effect of the maximum loop
    budget $K$ on \texttt{test-clean}, with 100h setup. Each point
    corresponds to a separately trained LARM model with the same shared
    encoder width ($d=384$).
    \subref{fig:checkpoint_interval_ablation} Effect of the checkpoint
    interval $c$, showing that moderately sparse supervision yields better
    results than both dense supervision and final-only supervision.
}
    \label{fig:loop_budget_and_supervision}
    \vspace{-0.5cm}
\end{figure}
\subsection{Ablation Study}

We ablate the main design choices of LARM on LibriSpeech 100h using
$d=384$ and report WER (\%) on \texttt{test-clean} with greedy CTC
decoding. Unless stated otherwise, the reference model uses $K=12$
loops, checkpoint interval $c=4$, FiLM depth conditioning,
previous-frame feedback, and learned mixing scalars $\alpha,\beta$.

\textbf{Naive looping.}
Before isolating individual components, we first compare LARM to a naive
looped encoder that reuses the same shared Transformer block for $K=12$
loops but removes the proposed loop-structuring mechanisms. Naive
recurrence already substantially improves over the non-looped 4-block
encoder, see Table~\ref{tab:ablation_naive_looping}, confirming that recurrent depth is useful. However, full LARM
further reduces WER from 12.70 to 11.34, showing that sparse
checkpointing, loop conditioning, and feedback make recurrent computation
more effective.

\textbf{Effect of trained loop budget.}
Figure~\ref{fig:loop_budget_and_supervision}(a) varies the maximum
trained loop budget $K$ across separately trained models, while keeping
the shared encoder width fixed at $d=384$. Increasing $K$ improves WER
from $K=1$ to $K=20$, showing that larger recurrent budgets can provide
additional refinement capacity. Performance degrades at $K=24$,
indicating that the shared recurrent dynamics have a useful operating
range and can saturate when the loop budget becomes too large. We also
observe that models trained with larger loop budgets can yield stronger
intermediate exits than models trained directly with shorter budgets,
suggesting that later supervised checkpoints may regularize earlier
recurrent states. Detailed per-loop exit results are reported in
Appendix~\ref{app:detailed_test_time_scaling}.

\textbf{Effect of checkpoint interval.}
Figure~\ref{fig:loop_budget_and_supervision}(b) varies the checkpoint
interval $c$, which controls how frequently CTC supervision is applied
within the recurrent loop. Dense supervision ($c=1$) is weaker than
moderately sparse supervision, suggesting that forcing every loop to act
as a complete recognition stage limits latent refinement. Final-only
supervision is also weaker, indicating that intermediate recognition
checkpoints are useful. Overall, the best results are obtained with
intermediate checkpoint intervals, supporting the use of supervised
recognition checkpoints separated by unsupervised refinement loops.

\textbf{Effect of depth conditioning.}
Table~\ref{tab:ablation_depth_conditioning} compares different ways of
conditioning the shared recurrent computation on loop depth. Removing
depth conditioning degrades performance, showing that the shared encoder
benefits from knowing its position in the recurrent process. Among the
tested variants, FiLM performs best, suggesting that feature-wise
rescaling and shifting provide a more effective mechanism for
stage-specific computation than additive embeddings or MLP-based
conditioning.

\textbf{Effect of recurrent feedback and aggregation.}
Table~\ref{tab:ablation_feedback} studies the recurrent feedback and
state aggregation mechanisms. Removing feedback substantially degrades
performance, showing that later loops benefit from access to previous
soft token hypotheses. Current-frame feedback already improves over no
feedback, while previous-frame feedback with learned mixing scalars gives
the best result. Learning $\alpha$ and $\beta$ is also important:
fixing these aggregation weights increases WER.

\begin{table}[t]
\centering
\caption{Ablation studies on LibriSpeech 100h. WER (\%) is reported on
\texttt{test-clean} with greedy CTC decoding.}
\label{tab:larm_ablation_components}
\renewcommand{\arraystretch}{1.08}
\small
\begin{subtable}[t]{0.25\linewidth}
\centering
\caption{Naive recurrence}
\label{tab:ablation_naive_looping}
\setlength{\tabcolsep}{4pt}
\begin{tabular}{lc}
\toprule
\textbf{Model} & \textbf{WER} \\
\midrule
4-block encoder & 26.78 \\
Naive looped & 12.70 \\
\rowcolor{larmbg}
LARM & \textbf{11.34} \\
\bottomrule
\end{tabular}
\end{subtable}
\hfill
\begin{subtable}[t]{0.25\linewidth}
\centering
\caption{Depth conditioning}
\label{tab:ablation_depth_conditioning}
\setlength{\tabcolsep}{4pt}
\begin{tabular}{lc}
\toprule
\textbf{Mode} & \textbf{WER} \\
\midrule
None & 12.13 \\
Embedding & 11.98 \\
MLP & 11.77 \\
\rowcolor{larmbg}
LARM & \textbf{11.34} \\
\bottomrule
\end{tabular}
\end{subtable}
\hfill
\begin{subtable}[t]{0.30\linewidth}
\centering
\caption{Feedback and aggregation}
\label{tab:ablation_feedback}
\setlength{\tabcolsep}{4pt}
\begin{tabular}{lc}
\toprule
\textbf{Variant} & \textbf{WER} \\
\midrule
Current & 11.43 \\
No feedback & 12.31 \\
fixed $\alpha,\beta$ & 11.91 \\
\rowcolor{larmbg}
LARM & \textbf{11.34} \\
\bottomrule
\end{tabular}
\end{subtable}
\vspace{-0.5cm}
\end{table}

\section{Conclusion}
We introduced LARM, a depth-conditioned looped Transformer that turns
recurrent encoder depth into a controllable test-time compute axis for
ASR. By combining sparse CTC checkpoints, supervision-clock embeddings,
FiLM depth conditioning, and delayed prediction feedback, LARM enables a
shared encoder to perform stage-specialized recurrent refinement.
Experiments on LibriSpeech show that LARM improves with additional
recurrent computation and achieves competitive WER against much deeper
unshared encoders while using substantially fewer parameters. These
results suggest that test-time compute scaling can extend beyond
autoregressive language modeling to continuous non-autoregressive speech
recognition.


\bibliographystyle{unsrt}
\bibliography{refs}


\appendix
\newpage

\section{Additional Architectural Details}
\label{app:architecture_details}

\subsection{CTC Prediction Head}
\label{app:ctc_head}

The CTC prediction head $\psi$ is implemented as a linear projection from
the hidden dimension $d$ to the output vocabulary $\mathcal{V}$:
\begin{equation}
    \boldsymbol{\ell}^{(k)}
    =
    \psi(\mathbf{z}^{(k)})
    =
    \mathbf{z}^{(k)} \mathbf{W}_{\psi}^{\top}
    +
    \mathbf{b}_{\psi},
    \qquad
    \mathbf{W}_{\psi} \in \mathbb{R}^{|\mathcal{V}| \times d}.
\end{equation}
Here $\boldsymbol{\ell}^{(k)} \in \mathbb{R}^{T' \times |\mathcal{V}|}$,
and $\mathcal{V}$ includes the CTC blank symbol. The corresponding token
posterior is
\begin{equation}
    \mathbf{p}^{(k)}
    =
    \mathrm{softmax}\!\left(\boldsymbol{\ell}^{(k)}\right).
\end{equation}

\subsection{Shared Encoder Details}
\label{app:shared_encoder}

The shared encoder $F_{\theta}$ consists of $N$ stacked pre-norm
Transformer blocks. Given an input representation $\mathbf{z}$, each
block applies multi-head self-attention (MHSA) followed by a feed-forward
network (FFN), both with residual connections:
\begin{equation}
    \mathbf{z}' = \mathbf{z} + \mathrm{MHSA}(\mathrm{LN}(\mathbf{z})),
    \qquad
    \mathbf{z}'' = \mathbf{z}' + \mathrm{FFN}(\mathrm{LN}(\mathbf{z}')).
\end{equation}

MHSA uses $H$ attention heads with per-head dimension $d_h=d/H$.
Queries, keys, and values are produced using a fused projection,
\begin{equation}
    [\mathbf{Q}, \mathbf{K}, \mathbf{V}] = \mathbf{z}\mathbf{W}_{QKV},
    \qquad
    \mathbf{W}_{QKV} \in \mathbb{R}^{d \times 3d},
\end{equation}
followed by splitting and reshaping across heads. Rotary positional
embeddings (RoPE) are applied to $\mathbf{Q}$ and $\mathbf{K}$ before
scaled dot-product attention:
\begin{equation}
    \mathrm{Attn}(\mathbf{Q}, \mathbf{K}, \mathbf{V})
    =
    \mathrm{softmax}\!\left(
    \frac{\mathbf{Q}\mathbf{K}^{\top}}{\sqrt{d_h}}
    \right)\mathbf{V}.
\end{equation}

The FFN is a two-layer MLP with GELU activation:
\begin{equation}
    \mathrm{FFN}(\mathbf{z})
    =
    \mathbf{W}_2\,\mathrm{GELU}(\mathbf{W}_1\mathbf{z} + \mathbf{b}_1)
    + \mathbf{b}_2,
\end{equation}
where $\mathbf{W}_1 \in \mathbb{R}^{4d \times d}$ and
$\mathbf{W}_2 \in \mathbb{R}^{d \times 4d}$.

The encoder $F_{\theta}$, CTC prediction head $\psi$, supervision clock
embedding projection $\mathbf{W}_c$, and FiLM conditioning models are shared
across all loops.

\begin{figure}[!h]
    \centering
    \includegraphics[width=0.7\linewidth]{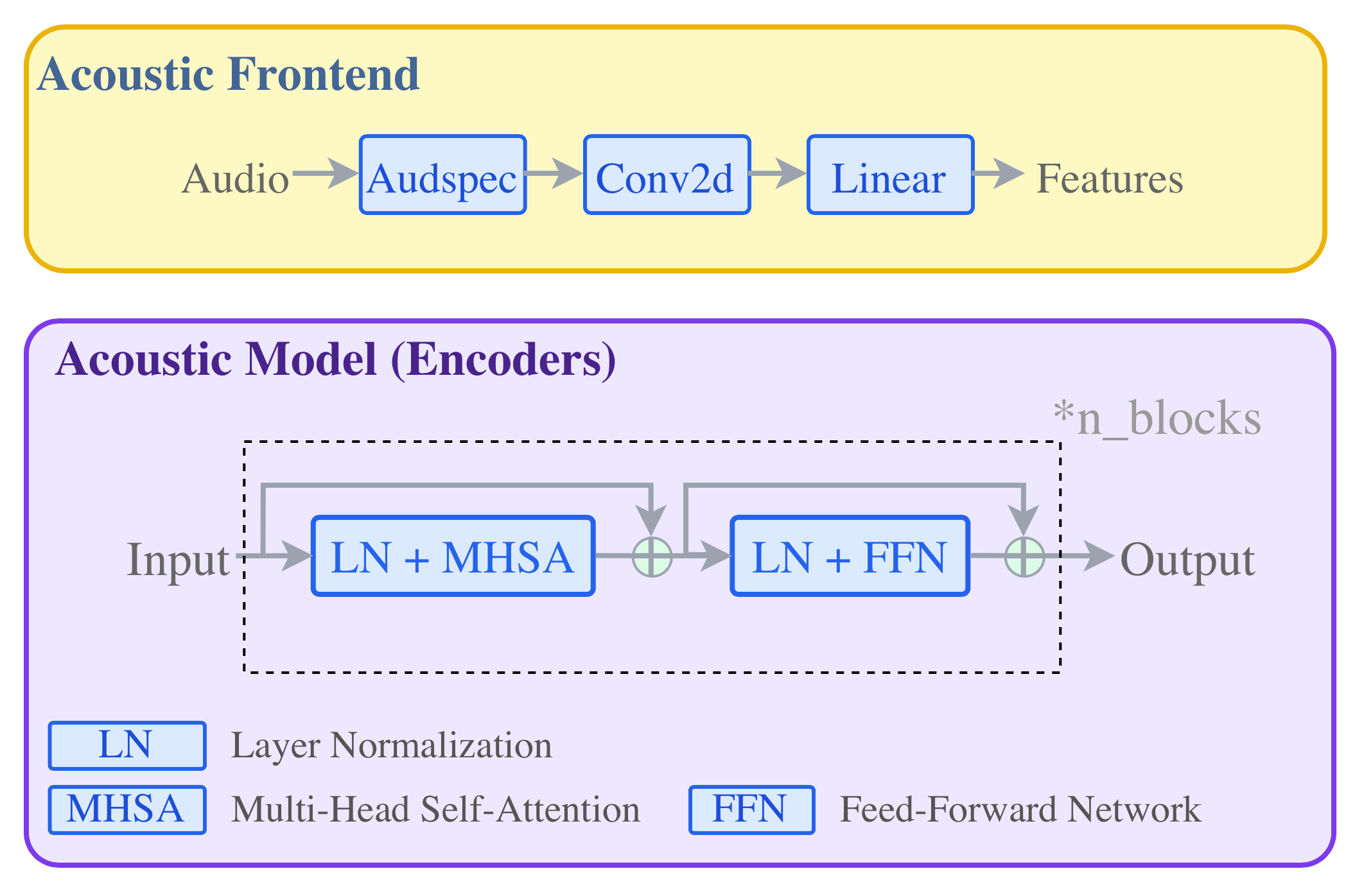}
\caption{
\textbf{Acoustic frontend and encoder block.}
The frontend converts audio features into latent representations, while
the encoder applies stacked pre-norm Transformer blocks with MHSA and FFN
modules.
}
    \label{fig:larm_architecture2}
\end{figure}
\section{Additional Experimental Setup}
\label{app:experimental_setup}

\paragraph{Input features.}
Audio is resampled to 16\,kHz and represented using 80-dimensional
log-Mel filterbank features, extracted with a 25\,ms window and a
10\,ms hop.

\paragraph{Output vocabulary.}
We use a character-level CTC vocabulary with 30 symbols: the 26
lower-case English letters, an apostrophe, a word-boundary symbol, a
\texttt{<unk>} symbol, and the CTC blank symbol.

\subsection{Experimental Architecture Details}
\label{app:experimental_architecture}

The convolutional frontend uses two strided 2D convolutional layers with
64 channels, kernel size $3\times3$, stride 2, padding 1, and SiLU
activations. This yields a $4\times$ time downsampling and reduces the
frequency axis from 80 to 20 bins. The flattened output
($64\times20=1280$ dimensions) is projected linearly to the model
dimension $d$, followed by an initial dropout of $0.1$.

The shared encoder consists of $N=4$ Transformer blocks with rotary
positional embeddings (RoPE, base $10{,}000$) and FFN expansion ratio 4.
All configurations use a fixed per-head dimension $d_h=64$, so the
number of attention heads is $H=d/d_h$. We experiment with model widths
$d\in\{384,768,1024, 1280\}$, corresponding to $H\in\{6,12, 16,20\}$ heads and FFN
hidden dimensions $\{1536,3072,4096,5120\}$.

Unless otherwise stated, the LARM loop runs for $K=12$ steps with
checkpoint interval $c=4$, yielding $|\mathcal{S}|=3$ supervised
checkpoints. The FiLM depth-conditioning network uses hidden dimension
64. Prediction feedback is computed in previous-frame mode. The mixing
scalars $\alpha$ and $\beta$ are initialized to $0.5$ and learned jointly.

\subsection{Training Details}
\label{app:training_details}

All models are optimized with AdamW~\citep{loshchilov2018decoupled} using
$\beta_1=0.9$, $\beta_2=0.999$, $\varepsilon=10^{-8}$, and weight decay
$5\times10^{-3}$. Gradient norms are clipped to $1.0$. The learning rate
follows a cosine decay schedule with linear warmup, peak learning rate
$\eta=7\times10^{-4}$, and minimum learning rate $0.03\,\eta$. The
warmup lasts 1{,}000 steps for 100\,h experiments and 10{,}000 steps for
960\,h experiments.

Most models are trained for 50 epochs, with extended-budget runs noted
separately. The effective batch size is 32 for 100\,h experiments and for
the main 960\,h scaling runs, using gradient accumulation when needed. 

SpecAugment~\citep{park2019specaugment} is applied throughout training
with a light policy: one frequency mask with $F_{\max}=15$ and two time
masks with time-mask ratio $p_T=0.02$. Masked regions are filled with the
utterance mean.

\section{Complementary Experimental Results}
\label{app:experiments}
\subsection{Detailed Test-Time Scaling Across Loop Exits}
\label{app:detailed_test_time_scaling}

Table~\ref{tab:loop_exit_scaling} reports WER at supervised loop exits
for LARM models trained with different maximum loop budgets on
LibriSpeech 100h. All models use checkpoint interval $c=4$, so supervised
exits occur every four recurrent loops. These results complement
Section~\ref{sec:test_time_compute_scaling} by showing how recognition
quality evolves across loop exits for several trained loop budgets.

Across $K=4$ to $K=20$, greedy WER generally improves from one supervised
checkpoint to the next, supporting the interpretation of additional loops
as useful recurrent refinement steps. The $K=24$ model is an exception:
although later checkpoints still improve over early ones, its final WER
is worse than shorter-loop models, consistent with the saturation behavior
observed in Figure~\ref{fig:loop_budget_and_supervision}.

The table also shows that training with a larger maximum loop budget can
improve intermediate exits. For example, the 8-loop exit of the $K=12$
model improves over the final exit of the $K=8$ model, and the 16-loop
exit of the $K=20$ model improves over the final exit of the $K=16$
model. This suggests that later supervised checkpoints can regularize
earlier recurrent states, although the effect weakens once the loop
budget becomes too large.

Language-model-assisted WER is less monotonic than greedy WER. In several
runs, the best LM-decoded WER occurs before the final loop exit. This
likely reflects an interaction between how the model handle intermediate CTC hypothesis (at supervised loop) and the LM decoding. We therefore use greedy decoding as the primary diagnostic
for recurrent refinement and report LM results for completeness.

\begin{table}[t]
\centering
\caption{Detailed loop-exit behavior on LibriSpeech 100h with checkpoint
interval $c=4$. WER (\%) is reported on \texttt{test-clean}. Greedy WER
is shown at supervised loop exits. LM columns report the best LM-decoded
WER across all loop exits and the final-loop LM WER.}
\label{tab:loop_exit_scaling}
\renewcommand{\arraystretch}{1.12}
\setlength{\tabcolsep}{4pt}
\resizebox{\linewidth}{!}{%
\begin{tabular}{c cccccc cc}
\toprule
\multirow{2}{*}{\textbf{Max loops $K$}}
& \multicolumn{6}{c}{\textbf{Greedy WER at supervised loop exits}}
& \multirow{2}{*}{\textbf{Best LM}}
& \multirow{2}{*}{\textbf{Final LM}} \\
\cmidrule(lr){2-7}
& \textbf{4} & \textbf{8} & \textbf{12} & \textbf{16} & \textbf{20} & \textbf{24}
& & \\
\midrule
4  & 14.07 & --    & --    & --    & --    & --    & 9.11 (4)  & 9.11 \\
8  & 14.22 & 12.07 & --    & --    & --    & --    & 8.67 (8)  & 8.67 \\
12 & 15.20 & 11.57 & 11.34 & --    & --    & --    & 8.38 (7)  & 8.66 \\
16 & 16.12 & 11.94 & 11.30 & 11.24 & --    & --    & 8.28 (7)  & 8.66 \\
20 & 16.95 & 11.83 & 11.17 & 11.05 & 11.02 & --    & 8.03 (7)  & 8.43 \\
24 & 20.05 & 14.33 & 13.52 & 13.39 & 13.31 & 13.34 & 9.02 (11) & 9.45 \\
\bottomrule
\end{tabular}%
}
\end{table}
\subsection{Extended Scaling Results}
\label{app:extended_results}

Table~\ref{tab:extended_scaling_results} reports additional larger-budget
runs on LibriSpeech 100h. These experiments are not intended as a single
controlled ablation, since they vary several axes across runs, but they
show the headroom available when increasing optimization budget,
recurrent loop budget, model width, or the number of shared Transformer
blocks. All results are reported on \texttt{test-clean}.

The clearest improvements come from combining longer training with a
larger loop budget. Compared with the main $d=768$, $K=12$, 50-epoch
result in Table~\ref{tab:larm_width_scaling} (9.58 greedy WER and 7.57
with LM), the $d=768$, $K=16$, 100-epoch run improves to 8.39 greedy WER
and 7.11 with LM. Similarly, compared with the main $d=1024$, $K=12$,
50-epoch result (8.95 greedy WER and 7.22 with LM), the $d=1024$,
$K=16$, 100-epoch run reaches 8.27 greedy WER and 6.95 with LM. These
results suggest that the main-table configurations do not saturate the
potential of LARM, and that larger recurrent budgets and longer
optimization remain beneficial.

Increasing width alone shows weaker returns at this scale. The
$d=1280$, $K=12$, 50-epoch run reaches 8.92 greedy WER, only slightly
better than the main $d=1024$, $K=12$ result, and its LM-assisted WER is
slightly worse. This is consistent with the saturation trend observed in
Section~\ref{sec:test_time_compute_scaling}: once the recurrent
configuration is fixed, simply increasing width provides diminishing
returns. In contrast, deeper shared encoders at small width
($N=6,K=8$ and $N=8,K=6$) improve over the default small $d=384$ model
from Table~\ref{tab:larm_width_scaling}, but remain behind the wider and
longer-trained LARM variants. This suggests that additional block depth
helps, but is less effective than jointly increasing width, recurrent
budget, and optimization time.

LM-assisted decoding is again less monotonic across loop exits. Several
extended runs obtain their best LM WER before the final loop, even when
greedy WER is best at the final loop. We therefore report both the final
LM result and the best LM result across loop exits. As discussed in Appendix~\ref{app:detailed_test_time_scaling}, we interpret greedy WER
as the cleaner diagnostic of recurrent acoustic refinement, because the interaction between
LM output and the intermediate hypothesis is complex and not explicitly controlled.

\begin{table}[t]
\centering
\caption{Extended LARM scaling results on LibriSpeech 100h
\texttt{test-clean}. WER (\%) is reported with greedy CTC decoding and
with a 4-gram language model. $N$ is the number of shared Transformer
blocks, $K$ is the maximum loop budget, and $c$ is the checkpoint
interval.}
\label{tab:extended_scaling_results}
\renewcommand{\arraystretch}{1.12}
\setlength{\tabcolsep}{5pt}
\resizebox{\linewidth}{!}{%
\begin{tabular}{ccccccccc}
\toprule
\textbf{$d$}
& \textbf{$N$}
& \textbf{$K$}
& \textbf{$c$}
& \textbf{Epochs}
& \textbf{Final greedy}
& \textbf{Best greedy}
& \textbf{Final LM}
& \textbf{Best LM} \\
\midrule
768  & 4 & 16 & 4 & 100 & 8.39 & 8.39 (16) & 7.11 & \textbf{6.83} (9) \\
768  & 4 & 16 & 8 & 100 & 8.50 & 8.50 (16) & 7.12 & 7.12 (16) \\
1024 & 4 & 16 & 4 & 100 & \textbf{8.27} & \textbf{8.27} (16) & \textbf{6.95} & 6.93 (12) \\
1280 & 4 & 12 & 4 & 50  & 8.92 & 8.92 (12) & 7.38 & 7.19 (8) \\
384  & 6 & 8  & 4 & 50  & 10.86 & 10.86 (8) & 8.25 & 8.25 (7--8) \\
384  & 8 & 6  & 3 & 50  & 10.85 & 10.85 (6) & 8.34 & 8.33 (4--5) \\
\rowcolor{larmbg}384  & 4 & 12  & 4 & 50  & 11.34 & 11.34 (12) & 8.66 & 8.38 (7) \\

\bottomrule
\end{tabular}%
}
\end{table}

\section{Broader Impact}
\label{app:broader_impact}

This work studies test-time compute scaling for ASR through recurrent
encoder depth. The main broader impact of LARM is that it changes how ASR
systems can be deployed under variable compute constraints. Instead of
training and maintaining separate shallow and deep encoders for different
latency or accuracy targets, a single shared-parameter model can expose
multiple operating points by changing the number of recurrent loops at
inference. This may be useful in settings where the available compute
budget changes across devices, users, or requests, while keeping the
stored model size fixed.

This also provides a different form of parameter efficiency from standard
model compression. LARM does not only reduce parameters relative to deep
unshared encoders; it makes additional computation available as a
runtime decision. Such a mechanism could support ASR systems that spend
more computation on difficult utterances and less on easier ones, although
the present work uses fixed loop budgets rather than a learned adaptive
policy.

The main deployment cost is that improved recognition is obtained by
spending more inference computation. Thus, LARM shifts part of the scaling
burden from parameters to latency and energy. This trade-off is central to
the method: the same mechanism that enables test-time compute scaling also
requires careful selection of loop budgets in practical systems. Future
adaptive early-exit mechanisms could make this trade-off more efficient by
allocating recurrent computation only when it improves recognition.

\section{Limitations}
\label{app:limitations}

The evaluation in this work is mainly limited to LibriSpeech. While this
provides a controlled benchmark for studying recurrent encoder depth, it
does not fully characterize the behavior of LARM on conversational speech,
accented speech, multilingual data, noisy recordings, far-field audio, or
overlapping speakers.

LARM currently uses fixed loop budgets. Although sparse CTC checkpoints
provide natural early-exit points, the model does not learn an
input-dependent halting rule. As a result, the number of loops must be
chosen externally, and all utterances are processed with the same maximum
budget unless an additional exit criterion is introduced.

Finally, LARM is implemented with CTC prediction feedback and evaluated
primarily with greedy CTC decoding, with fixed 4-gram LM decoding used for
the main benchmark results. The interaction between recurrent acoustic
refinement and stronger decoding methods remains
open. Extending the proposed loop conditioning and delayed feedback
mechanisms to transducer or attention-based ASR models is left for future
work.



\end{document}